\documentclass[11pt,a4paper]{article}
\usepackage{times,latexsym, amssymb, graphicx, amsmath}
\usepackage{url}
\usepackage[T1]{fontenc}

\usepackage{tikz}
\usetikzlibrary{shapes.geometric, arrows, calc, positioning, shapes, fit}
\usepackage{algorithm2e}
\RestyleAlgo{ruled}
\usepackage{listings}
\usepackage{multirow}
\usepackage{pgffor} %
\usepackage{booktabs}
\usepackage{cprotect}
\usepackage[table]{xcolor}
\usepackage{float}
\usepackage{array}
\usepackage{amsmath}
\usepackage{tabularx, booktabs}
\usepackage{xcolor}
\usepackage{soul}
\providecommand{\dev}[1]{#1} %
\definecolor{lightpink}{RGB}{255,209,220}
\DeclareRobustCommand{\pk}[1]{#1} %
\soulregister{\emph}{1} %
\usepackage{subcaption}
\newcolumntype{L}[1]{>{\raggedright\arraybackslash}p{#1}}

\usepackage[acceptedWithA]{tacl2021v1}

\usepackage{xspace,mfirstuc,tabulary}

\newif\iftaclinstructions
\taclinstructionsfalse %
\iftaclinstructions
\renewcommand{\confidential}{}
\renewcommand{\anonsubtext}{(No author info supplied here, for consistency with
TACL-submission anonymization requirements)}
\newcommand{\instr}
\fi

\iftaclpubformat %

\else

\fi

\title{Semantic Needles in Document Haystacks: Sensitivity Testing of LLM-as-a-Judge Similarity Scoring}

\author{
  \textbf{Sinan G.\ Aksoy\textsuperscript{1}},
  \textbf{Alexandra A.\ Sabrio\textsuperscript{1,2}},
  \textbf{Erik VonKaenel\textsuperscript{1,3}},
  \textbf{Lee Burke\textsuperscript{1}} \\
  \textsuperscript{1}Pacific Northwest National Laboratory \\
  \textsuperscript{2}Washington University in St.\ Louis \\
  \textsuperscript{3}Humana Inc.\ \\
  \texttt{\{first.last\}@pnnl.gov},
  \texttt{vonkaenelerik@gmail.com}
}

\date{}

\begin{document}
\maketitle

\begin{abstract}
We propose a scalable, multifactorial experimental framework that systematically probes LLM sensitivity to subtle semantic changes in pairwise document comparison. We analogize this as a needle-in-a-haystack problem:
a single semantically altered sentence (the needle) is embedded within surrounding context (the hay), and we vary the perturbation type (negation, conjunction swap, named entity replacement), context type (original vs.\ topically unrelated), needle position, and document length across all combinations, testing \pk{seven} LLMs on tens of thousands of document pairs. 
Our analysis reveals several striking findings. First, LLMs \pk{can} exhibit a within-document positional bias distinct from previously studied candidate-order effects: \pk{some} models penalize semantic differences more harshly when they occur earlier in a document. 
Second, when the altered sentence is surrounded by topically unrelated context, it \pk{raises similarity scores in almost all settings}.
This is consistent with a \pk{contradiction account, in which topically related context allows the alteration to conflict with the surrounding sentences}.
\pk{Third, all models exhibit the same hierarchy in how harshly they score different perturbation types, with negation penalized the most, but express this hierarchy on significantly different scales. Finally, varying the prompt's conception of semantic similarity across several alternatives changes the magnitude but not the direction of these effects.} 
Together, these results demonstrate \pk{how} LLM semantic similarity scores are sensitive to \pk{the aforementioned parameters} in ways that go beyond the semantic change itself\pk{; our} framework offers a practical toolkit for auditing and comparing scoring behavior across current and future models.
\end{abstract}

\section{Introduction}

LLM-as-a-judge systems are rapidly pervading domains ranging from complex scientific workflows to routine, everyday decisions.
In these systems, an LLM acts as an automated evaluator: assessing the quality of generated text \citep{liu2023geval,zheng2023judging}, \pk{evaluating translation quality \mbox{\citep{Kocmi2023}},} grading student essays \citep{Song2024}, scoring medical question-answering \citep{Krolik2024}, evaluating mathematical reasoning \citep{Li2024}, and serving as a stand-in for costly human annotation across natural language generation tasks \citep{Chiang2023,dubois2023alpacafarm}.
Recent surveys document the remarkable breadth and rapid adoption of this paradigm \citep{Gu2025}, while simultaneously raising concerns about systematic biases in LLM evaluation \citep{wang2023large,Chen2024}.
As new LLMs are released at an accelerating pace and these judge systems are continually updated, practitioners face a moving target: each model version may introduce new scoring behaviors, biases, or idiosyncrasies that are difficult to anticipate.
For these reasons, it is increasingly important that we have experimental pipelines for evaluating LLM-as-a-judge systems that are \textit{detailed}, \textit{generalizable}, and \textit{scalable}.

One particularly far-reaching use case of LLM-as-a-judge systems is \textit{pairwise document similarity assessment}.
This task arises across a wide range of applications, including plagiarism and near-duplicate detection, information retrieval and document clustering, semantic textual similarity benchmarks \citep{cer2017semeval}, and automated scoring of genre-specific articles. %
Beyond settings where document comparison is the primary objective, pairwise similarity assessment is a critical subcomponent of evaluating \textit{other} tools.
For example, when testing text privatization, pairwise document similarity can quantify how much privatized text strays from the original semantic meaning \pk{\mbox{\citep{Staab2024}}}. 
Furthermore, the pairwise case can be naturally bootstrapped to multi-way comparisons, making robust pairwise evaluation all the more consequential.

Within this domain, the ability to detect and differentiate between \textit{subtle} semantic changes is becoming especially important.
There are several reasons for this.
First, small semantic differences are critical to many real-world applications of pairwise document similarity, beyond the aforementioned example of text privatization. 
In medical documentation, even minor revisions to treatment recommendations or dosage instructions can lead to substantially different patient outcomes, making detection of subtle semantic drift a matter of safety.
Similarly, in legal texts, small changes in wording, such as the substitution of ``and'' for ``or'' in a contractual clause, can dramatically alter the obligations and rights of the parties involved \citep{adams2006revisiting}.

Second, as LLMs become increasingly capable at pairwise similarity assessment, we need correspondingly challenging tasks to stress-test them and expose their limitations. Idiosyncratic scoring behaviors are often most visible ``on the margins,'' where subtle differences push the boundary of what a model can reliably detect \citep{Wang2023}. This makes fine-grained tasks more discriminative than coarser evaluations.

Third, even humans are unreliable at detecting subtle semantic changes, raising the question of whether LLMs inherit similar blind spots or introduce new ones.
Research on semantic illusions suggests readers \pk{can} routinely overlook meaning alterations embedded within coherent discourse \citep{Cook,Nieuwland}.
Understanding whether LLMs exhibit analogous biases (or different ones) when scoring document similarity is essential for calibrating trust in these systems and for identifying failure modes before they propagate into downstream applications.

Accordingly, in this work we propose, apply, and analyze an experimental pipeline for testing LLM-as-a-judge sensitivity to semantic similarity between pairs of documents.
\pk{In the spirit of behavioral testing \mbox{\citep{Ribeiro2020}}, the pipeline characterizes how a judge's scores respond to controlled changes in its input, rather than whether those scores are correct.}
We analogize our experiment as a \textit{needle-in-a-haystack} problem.
While this metaphor originates in work on long-context retrieval (where a planted fact must be located within filler text \citep{kamradt2023needle,hsieh2024ruler}) we invoke it here to instead test how LLMs \textit{score semantic similarity} between two almost-identical documents, one of which contains a single semantic alteration embedded within surrounding context.
In our formulation, the ``needle'' is a single, semantically altered sentence, and the ``hay'' is the surrounding context of varying length, position, and relevance.
By systematically varying the needle type, hay type, hay amount, and needle position across all possible combinations and testing on tens of thousands of document pairs, we construct a scalable experimental design that probes LLM scoring sensitivity along multiple axes simultaneously. Our main contributions are two-fold:

 \begin{itemize}
      \item We propose an LLM-agnostic, automated, and scalable factorial experimental design for
  sensitivity testing of LLM-guided pairwise document semantic similarity.
      \item We instantiate this framework across seven LLMs\pk{, spanning several axes of comparison: variants within a model family (three GPT 5.6 models), vendors (OpenAI, Anthropic, Google), open (Gemma 4) vs.~closed weights, and model generations within a vendor (GPT-4o vs.~5.6). We develop} quantitative analyses, including positionality bias
  measures, document length effects, and \pk{comparisons across scoring prompts},
  that expose interpretable differences in scoring behavior.
  \end{itemize}

\begin{figure*}[t]
\centering
\small
\setlength{\fboxsep}{1.5pt} %

\begin{tikzpicture}[font=\small]

\node[draw=gray!50, rounded corners=6pt, fill=gray!5, inner sep=5pt] (main) {
\begin{minipage}{0.97\textwidth}

    \begin{minipage}[t]{0.58\textwidth}
        \centering
        \textbf{Original Document $d$}\\[-2pt]
        {\color{gray!60}\rule{0.6\textwidth}{0.4pt}}\\[4pt]
        \colorbox{cyan!15}{$\dots$}\\[1pt]
        \colorbox{cyan!15}{Thousands of flights land in Chicago.}\\[3pt]
        \colorbox{cyan!15}{Tons of passengers transfer at O'Hare.}\\[3pt]
        \colorbox{yellow!30}{O'Hare is larger and busier than Midway.}\\[3pt]
        \colorbox{cyan!15}{Airport runways handle traffic all day.}\\[3pt]
        \colorbox{cyan!15}{United runs its busiest hub at O'Hare.}\\[3pt]
        \colorbox{cyan!15}{$\dots$}
    \end{minipage}
    \hfill
    \begin{minipage}[t]{0.38\textwidth}
        \centering
        \textbf{Legend}\\[-2pt]
        {\color{gray!60}\rule{0.8\textwidth}{0.4pt}}\\[2pt]
        \footnotesize
        \begin{tabular}{@{}l@{}} 
            \colorbox{yellow!30}{\phantom{x}} Needle sentence \\[1pt]
            \colorbox{cyan!15}{\phantom{x}} \texttt{orig}: original hay \\[1pt]
            \colorbox{magenta!12}{\phantom{x}} \texttt{rand}: random hay \\[1pt]
            \textbf{\textcolor{red!70!black}{\underline{Underline}}} Semantic change \\[1pt]
            \texttt{neg}: Negation insertion \\[1pt]
            \texttt{con}: Conjunctive swap \\[1pt]
            \texttt{ner}: Named entity replacement \\[1pt]
            $(i,j)$: $i$ pre \& $j$ post sentences
        \end{tabular}
    \end{minipage}

    \vspace{0.2em}
    \centering \textbf{Example Comparisons}\\[-2pt]
    {\color{gray!60}\rule{0.4\textwidth}{0.4pt}}
    \vspace{0.2em}

    \centering
    \setlength{\tabcolsep}{0pt}
    \begin{tabular}{ m{6.6cm} @{\hspace{-1mm}} c @{\hspace{6mm}} m{6.6cm} @{\hspace{3mm}} c }
        
        \centering \footnotesize\color{gray!60} $d(\varnothing, \texttt{orig}, (1,2))$ & & 
        \centering \footnotesize\color{gray!60} $d(\texttt{neg}, \texttt{orig}, (1,2))$ & \\[2pt]
        
        \raggedright 
        \colorbox{cyan!15}{Tons of passengers transfer at O'Hare.}\\[1pt]
        \colorbox{yellow!30}{O'Hare is larger and busier than Midway.}\\[1pt]
        \colorbox{cyan!15}{Airport runways handle traffic all day.}\\[1pt]
        \colorbox{cyan!15}{United runs its busiest hub at O'Hare.}
        & \large\bfseries\color{gray!30} vs 
        & \raggedright 
        \colorbox{cyan!15}{Tons of passengers transfer at O'Hare.}\\[1pt]
        \colorbox{yellow!30}{O'Hare is \textcolor{red!70!black}{\textbf{\underline{not}}} larger and busier than Midway.}\\[1pt]
        \colorbox{cyan!15}{Airport runways handle traffic all day.}\\[1pt]
        \colorbox{cyan!15}{United runs its busiest hub at O'Hare.}
        & \shortstack{\footnotesize\textbf{Score}\\ \footnotesize 77/100} \\
        
        \multicolumn{4}{c}{\rule{0pt}{1.5ex} {\color{gray!20}\rule{0.98\linewidth}{0.4pt}} \rule{0pt}{1.5ex}} \\

        \centering \footnotesize\color{gray!60} $d(\varnothing, \texttt{rand}, (2,0))$ & & 
        \centering \footnotesize\color{gray!60} $d(\texttt{con}, \texttt{rand}, (2,0))$ & \\[2pt]

        \raggedright 
        \colorbox{magenta!12}{Goodfellas was a hit film in 1990.}\\[1pt]
        \colorbox{magenta!12}{Robert De Niro's acting was praised.}\\[1pt]
        \colorbox{yellow!30}{O'Hare is larger and busier than Midway.}
        & \large\bfseries\color{gray!30} vs 
        & \raggedright 
        \colorbox{magenta!12}{Goodfellas was a hit film in 1990.}\\[1pt]
        \colorbox{magenta!12}{Robert De Niro's acting was praised.}\\[1pt]
        \colorbox{yellow!30}{O'Hare is larger \textcolor{red!70!black}{\textbf{\underline{or}}} busier than Midway.}
        & \shortstack{\footnotesize\textbf{Score}\\ \footnotesize 89/100} \\

        \multicolumn{4}{c}{\rule{0pt}{1.5ex} {\color{gray!20}\rule{0.98\linewidth}{0.4pt}} \rule{0pt}{1.5ex}} \\

        \centering \footnotesize\color{gray!60} $d(\varnothing, \texttt{orig}, (0,1))$ & & 
        \centering \footnotesize\color{gray!60} $d(\texttt{ner}, \texttt{orig}, (0,1))$ & \\[2pt]

        \raggedright 
        \colorbox{yellow!30}{O'Hare is larger and busier than Midway.}\\[1pt]
        \colorbox{cyan!15}{Airport runways handle traffic all day.}
        & \large\bfseries\color{gray!30} vs 
        & \raggedright 
        \colorbox{yellow!30}{O'Hare is larger and busier than \textcolor{red!70!black}{\textbf{\underline{Chicago}}}.}\\[1pt]
        \colorbox{cyan!15}{Airport runways handle traffic all day.}
        & \shortstack{\footnotesize\textbf{Score}\\ \footnotesize 62/100} \\

    \end{tabular}
\end{minipage}
};
\end{tikzpicture}
\caption{An example illustrating our experimental pipeline. %
}
\label{fig:d-variants}
\end{figure*}

\noindent Our framework does not make value judgments about LLM scoring behavior, nor does it rank model performance.
Rather, it provides a modular experimental framework, within the familiar needle-in-a-haystack metaphor, through which one can generate rich experimental data about LLM scoring behavior and idiosyncrasies, for both current and future systems.
Our instantiation and subsequent analysis on real data demonstrate that the framework is highly discriminative, sharply highlighting differences between different LLMs as well as between versions of the same LLM.
Together, the intuitive needle-and-hay framing that practitioners can readily adapt to new perturbation types, domains, and models, and the quantitative analyses we develop for interpreting the resulting data offer a practical toolkit for anyone seeking to understand, compare, or audit LLM scoring behavior. \pk{Before proceeding, we briefly clarify how our work fits within the broader landscape of LLM-as-a-judge, and also position ourselves within related work.}

\paragraph{Relationship to LLM-as-a-judge literature}
\pk{We study the reference-based form of LLM-as-a-judge, in which a judge scores how well a text preserves the meaning of a reference, which is one of three judging variants identified by \mbox{\citet{zheng2023judging}}. Whereas LLM-as-a-judge studies most often focus on testing whether such judges are \emph{accurate}, we focus on how their verdict for fixed, atomic semantic changes responds to other parameters like where the change occurs, what surrounds it, and how the prompt defines semantic similarity. Studies like ours are sometimes referred to as ``behavioral testing'' of LLM systems} \citep{Ribeiro2020,Hu2024,Mahbub2026}.

\paragraph{Related work on needle-in-a-haystack probes}
\pk{A needle-in-a-haystack paradigm has appeared in multiple prior studies, though for varying purposes. The most popular focus is on testing \emph{retrieval} from long filler text} \citep{kamradt2023needle,hsieh2024ruler}\pk{. Work in this vein has shown that LLM performance depends on where relevant content appears \mbox{\citep{Liu2023}} and on the length, type, and location of surrounding padding \mbox{\citep{Levy2024}}. Closer to our setting, \mbox{\citet{Kurfali2025}} embed mutually contradictory needles while varying their position and the relevance of the haystack, \mbox{\citet{LiContraDoc2024}} plant self-contradictions, such as negations and altered entity attributes, in documents of varying length, and \mbox{\citet{Fu2025}} give models an original document alongside a copy from which content has been removed. Crucially, these studies focus on scoring retrieval, question answering, or detection. \emph{In short, typical needle-in-a-haystack tests ask whether a model can find the needle, whereas we hand the judge the needle in plain sight and ask how much it counts.}}

\paragraph{Related work on behavioral testing}
\pk{Behavioral testing evaluates a model through controlled changes to its input rather than aggregate benchmark accuracy \mbox{\citep{Ribeiro2020}}. Perturbation tests of this kind have exposed blind spots of learned evaluation metrics} \citep{Sai2021,He2023} \pk{and of LLM evaluators, which failed to detect targeted errors \mbox{\citep{Doddapaneni2024}}, and which confuse distinct quality criteria \mbox{\citep{Hu2024}}. Positional effects have been documented for embedding models} \citep{Coelho2024,Lee2024}\pk{, and LLM-based faithfulness metrics have been shown to be sensitive to the order of the source document \mbox{\citep{Wan2025}}. \emph{Whereas other behavioral tests check whether and how a judge reacts to meaningful changes \mbox{\citep{Mahbub2026}}, we instead test a judge's response to change in the presence of moderators such as where that change occurs and what surrounds it---and how all of these interact within a factorial design.}}

Our work is organized as follows: Section~\ref{sec:experimental-overview} describes our experimental design, Section~\ref{sec:results} presents our results, and Section~\ref{sec:conc} discusses our findings in context.

\section{Experimental Design}
\label{sec:experimental-overview}

\paragraph{Overview} %
We study how LLMs quantitatively score semantic similarity between pairs of almost-identical text documents apart from a {\it needle} -- a  single, semantically altered sentence. We consider three needle types: semantic perturbation via negation insertion, conjunction replacement, and named entity replacement. These semantically altered sentences are subsequently positioned between some number of preceding and succeeding sentences -- the surrounding {\it hay}. We vary both the relative {\it position} of the needle within this hay (moving the semantic difference across beginning, middle, and end) and the total hay amount (the length of the document by sentence count). We also toggle the hay type; if the sentences surrounding the needle are taken from a randomly chosen document, we call this ``random hay'', whereas ``original hay'' retains the original surrounding context. %
We vary all possible combinations of these parameters, automate needle and hay insertion, run each setting on many documents, and repeat this across multiple LLMs. %

\paragraph{Formal description}
Let $\mathcal{U}$ denote a universe of raw text documents. We take $\mathcal{U}$ to be a subset of \texttt{Plain Text Wikipedia} \cite{plain_text_wikipedia_2020} consisting of 453,602 documents. We further clean these to remove non-natural language text, and filter them for length, yielding 40,003 cleaned documents, $\mathcal{C}$. There are four main parameters to our experiment:

\begin{itemize}
    \item Needle type: $N\in \{\varnothing, \texttt{neg},\texttt{con}, \texttt{ner}\}$, the semantic change type applied to the altered sentence. Can be negation (\texttt{neg}), swapping between ``and'' and ``or'' (\texttt{con}), swapping a named entity (\texttt{ner}), or no change ($\varnothing$).
    \item Hay type: $H \in \{\texttt{orig},\texttt{rand}$\}, the type of sentences preceding and following the altered sentence. Can be the original document's sentences (\texttt{orig}) or consecutive sentences from a random document in $\mathcal{C}$ (\texttt{rand}). 
    \item Position: $P\in \{(i,j) : i,j \in \{0,\dots,k\}\}$, the number of sentences $i$ preceding and number of sentences $j$ following the altered sentence, where $i$ and $j$ are integers between $0$ and $k$. We take $k=9$.
    \item LLM: $L \in$ \{GPT-4o, GPT 5.6 Luna, GPT 5.6 Terra, GPT 5.6 Sol, Claude 5 Sonnet, Gemini 3.7 Flash, Gemma 4\}, the Large Language Model scoring document similarity.
\end{itemize}

We consider each possible parameter setting $(L,N,H,P)$ in the Cartesian product
\[
\mbox{LLM} \times (\mbox{Needle} \setminus \{\varnothing\}) \times \mbox{Hay} \times \mbox{Position},
\]
which, for our choices detailed above, totals \pk{4200} distinct parameter settings. For each position $(i,j) \in P$, we randomly permute the documents in $\mathcal{C}$ and process them in this order so that the set of documents processed by an LLM, needle, and hay triple is the same random sample. %
Fixing a position $(i,j)$, we then run the following pipeline multiple times until a stopping criterion is met: for a document $d \in \mathcal{C}$ with $|d|$ many sentences, we select $m=\lceil\frac{|d|}{2} \rceil$ as the ``middle'' sentence to be semantically altered. The resulting document, $d(N,H,P)$, is derived from $d$ with $i+j+1$ many sentences, where the $m\textsuperscript{th}$ sentence of $d$ has been altered via semantic change $N$, and this sentence is preceded and followed by $i$ and $j$ many sentences of type $H$, respectively. We then use LLM $L$ to compare this document with its unaltered counterpart, i.e.
 \[
 d(\varnothing, H,P) \mbox{  vs.  } d(N,H,P)
 \]
using a basic scoring prompt for semantic similarity (see Figure \ref{fig:criteria}, Appendix \ref{sec:appendix_params}). The resulting score, $s(N,H,P)$, is an integer in $[0,100]$. Figure \ref{fig:d-variants} illustrates the comparison for different triples in $N \times H \times P$ on a toy example. We continue scoring documents until at least 100 documents have been processed and a stopping criterion based on mean score stabilization is also met. Finally, for a fixed $(L,N,H)$, we take the maximum number of documents run across all positions, and run all other positions on this number of documents. %
For more details on data cleaning, needle implementation, and our stopping criteria, see Appendix \ref{sec:appendix_params}.

\section{Results}\label{sec:results}
\paragraph{Positionality}

We begin by investigating how the {\it position} of the perturbed sentence within the document affects similarity scores. %
Do LLMs penalize semantic differences more harshly when they appear early in a document, and how rapidly does this bias intensify with position? Does surrounding context that is topically unrelated to the altered sentence amplify or dampen this effect?
To begin, a simple null hypothesis is that scores are identical (in distribution) whether the needle is in the first vs second half of the document, i.e.  
\[
\{s(N,H, (i,j)): i>j\} = \{s(N,H,(i,j)): i<j\}
\]

\begin{figure}[t]
    \centering
    \includegraphics[width=\linewidth]{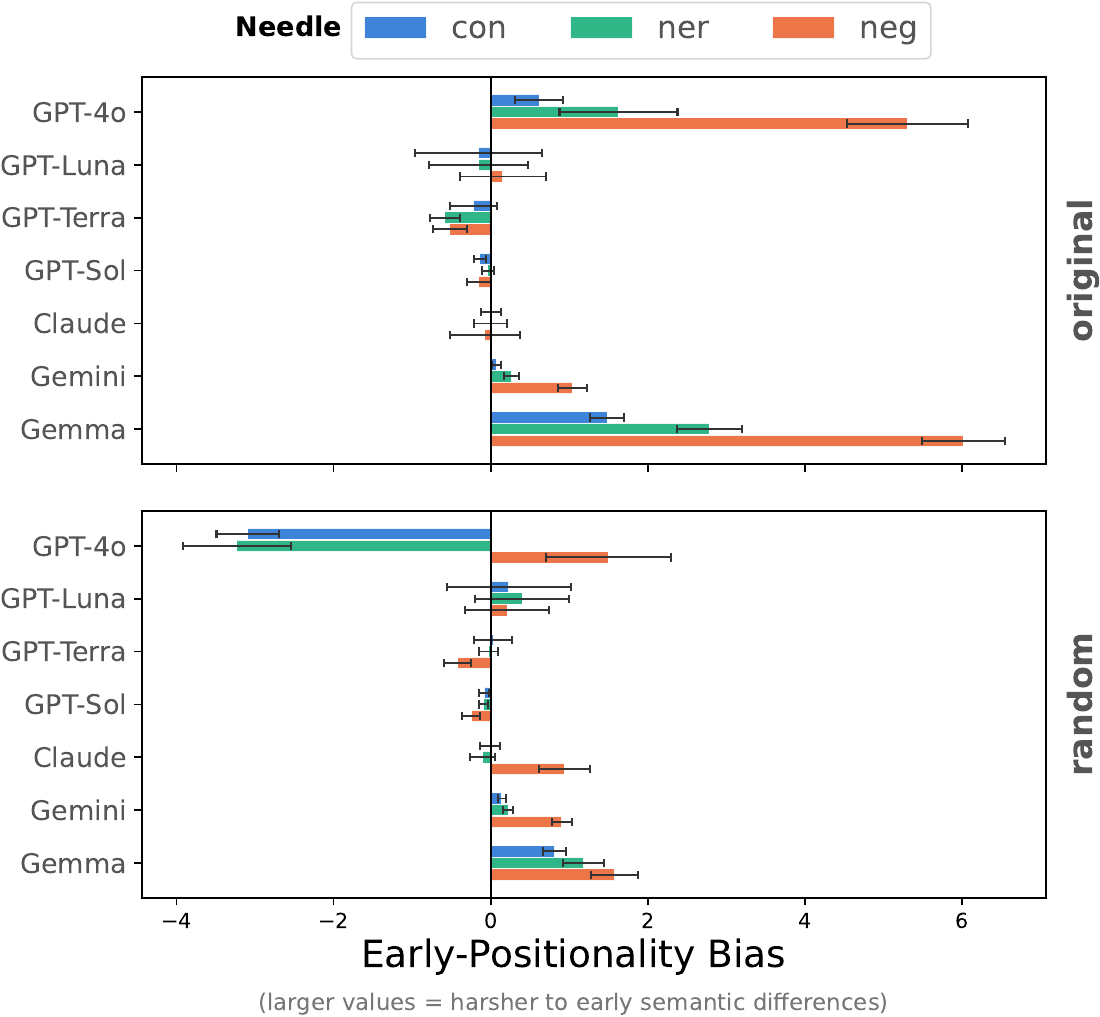}
    \caption{The early-positionality bias \pk{(with 95\% CIs)} for different LLMs and needle types under original (top) and random (bottom) hay.}
    \label{fig:epb}
\end{figure}

We perform a 2-sample Kolmogorov-Smirnov test between the score distributions of these ``first-half'' vs ``second-half'' needle documents for each LLM, needle, and hay parameter setting, for a total of \pk{42} hypothesis tests. \pk{After Bonferroni-Holm correction, we reject this hypothesis at $p<0.05$ in 31 of the 42 settings, with most exceptions from Claude and GPT-Sol. While positionality thus matters for the other LLMs, on a macro scale the differences are small for most of them, with two notable exceptions. To quantify this, we compute the mean score when the needle is in the second half minus the mean score when it is in the first half, so that positive values indicate harsher penalization of earlier semantic differences. Accordingly, we call this the Early-Positionality Bias, and plot it for each setting in Figure~\mbox{\ref{fig:epb}}.}

\pk{Under original hay (top), the bias is at most about one point, except for GPT-4o and Gemma, which penalize early needles by up to 5.3 and 6.0 points, respectively. For both, as for Gemini's smaller bias, the effect is largest for \texttt{neg}. Random hay (bottom) shrinks Gemma's bias and GPT-4o's \texttt{neg} bias, and reverses the direction of GPT-4o's bias for \texttt{con} and \texttt{ner}. Its effect is not uniform, however: Claude, unbiased under original hay, acquires a modest \texttt{neg} bias (0.9 points). In short, sizable early-positionality bias is exhibited by select LLMs, and is sensitive to hay type.}

\begin{figure*}[t]
    \centering
        \includegraphics[width=\linewidth]{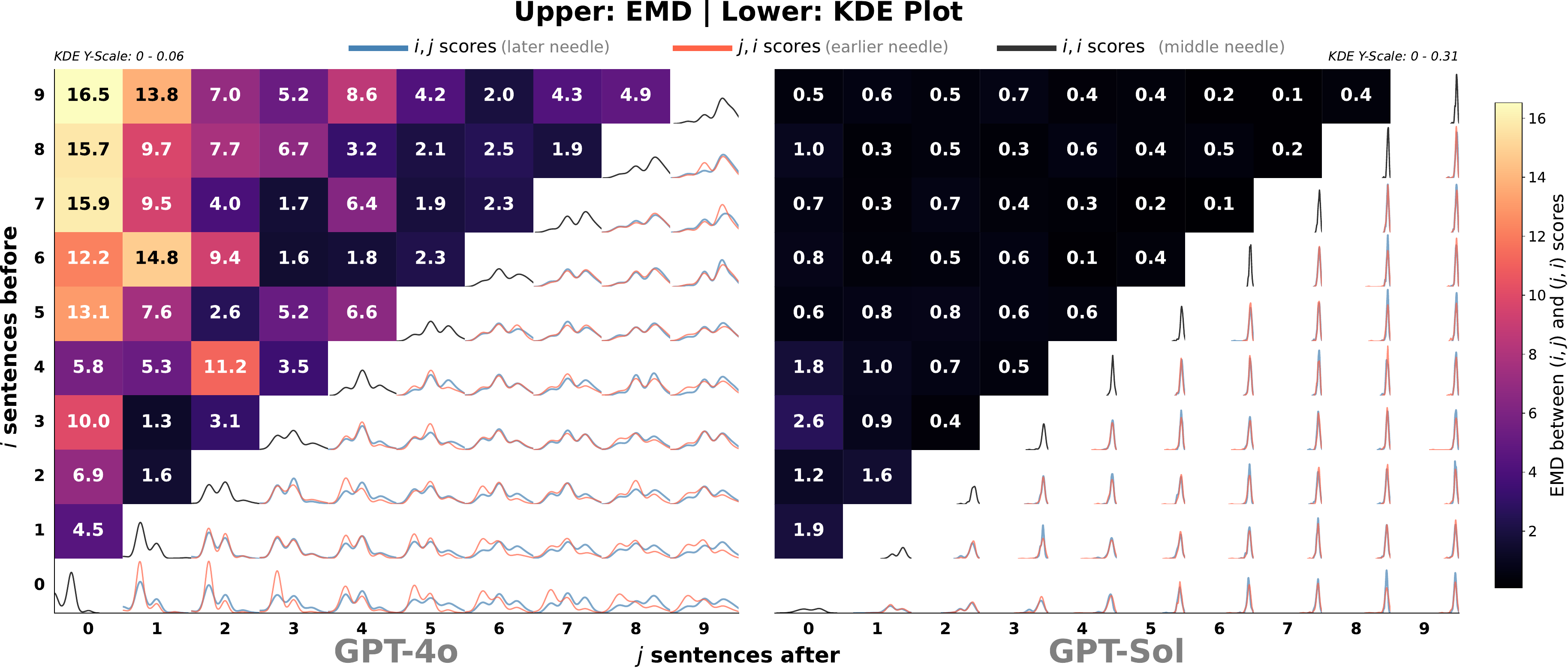}
        \caption{EMD and KDE plot comparison of score distributions for GPT-4o (left) and \pk{GPT-Sol} (right) by $i,j$ position, under the \pk{\texttt{neg}} needle and \texttt{orig} hay. Note the y-axis scale difference between panels. %
        } \label{fig:hybrid_heatmap}
\end{figure*}

\pk{These inter-LLM differences become starker in a finer-grained, position-by-position analysis.} For example, Figure \ref{fig:hybrid_heatmap} compares GPT-4o and \pk{GPT-Sol} for original hay, and the \pk{\texttt{neg}} needle type. Each $(i,j)$ cell in the upper portion displays the earth mover's distance (EMD) between score distributions for $(i,j)$ and $(j,i)$. One takeaway is the brightly colored first column which indicates high EMD: GPT-4o's scores more sharply differentiate documents with perturbed sentences that appear first vs.~last in the document\pk{, by up to 16.5 points}. \pk{GPT-Sol, by contrast, shows almost no positionality bias, with EMD below 3 in every cell.}

The lower portion of each heatmap contains Kernel Density Estimation (KDE) plots of the distributions in question, such that the $(j,i)$ cell plots the distributions compared by EMD in the $(i,j)$ cell. %
For example, the high EMD in GPT-4o's left column can also be seen by the divergence of the red and blue lines in the bottom row. Studying the shape of these plots as we vary $i$ and $j$ yields nuanced differences within and across LLMs. For example, with larger $i,j$, \pk{GPT-Sol's} distribution becomes sharply concentrated at high-scores; the $(9,9)$ cell shows consistent, near perfect similarity scores for 19-sentence documents differing in their middle sentence. In contrast, GPT-4o's scores remain more spread out for this cell and appear \pk{multimodal} throughout\pk{, peaking at the boundaries of the rubric's scoring bands (Figure~\mbox{\ref{fig:criteria}}): 78\% of its scores are 25, 26, 50, 51, 75, or 76}. These observations suggest LLMs\pk{, even two GPT models,} exhibit qualitatively different scoring ``fingerprints'' based on needle position.

\paragraph{Document length}

We now explore how document length affects scoring behavior, regardless of needle position. Measured in sentences, the length of the document for position $(i,j)$ is simply $k=i+j+1$. As $k$ increases, the proportion of the document's sentences that have been semantically altered decreases. Accordingly, a natural hypothesis is that semantic similarity score is increasing in document length and proportional to the fraction of semantically identical sentences, i.e. $\frac{k-1}{k}$.

\begin{figure}[t]
    \centering
    \includegraphics[width=\linewidth]{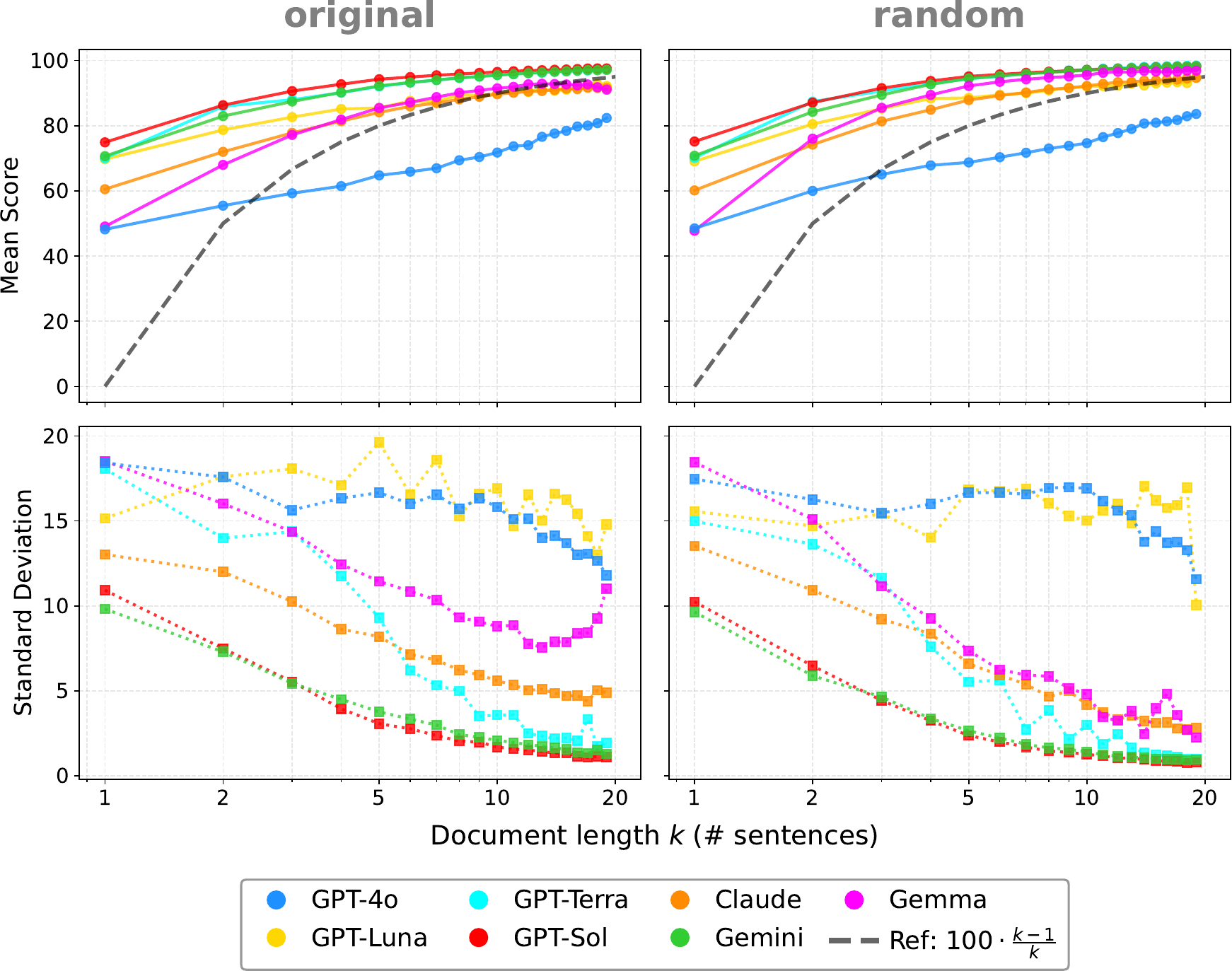}
    \caption{The mean (top) and standard deviation (bottom) of scores by document length for different LLMs\pk{, averaged over the three needle types,} under either \texttt{orig} or \texttt{rand} hay.}
    \label{fig:doc_length}
\end{figure}

Figure \ref{fig:doc_length} plots the mean and standard deviation of scores\pk{, averaged over the three needle types}. The mean score plots include a reference line of $100\cdot \tfrac{k-1}{k}$. Scores above this line suggest semantic perturbation preserves {\it some} of the semantic meaning of the altered sentence, while scores below suggest semantic perturbation alters the meaning of the document more severely than proportionally to the number of altered sentences. For both \texttt{orig} and \texttt{rand} hay, all LLMs have mean scores that increase in document length, as hypothesized. \pk{For \texttt{orig} hay, only GPT-Terra, GPT-Sol, and Gemini stay above the reference line at every length, while GPT-4o falls below it at nearly every length.} 

\pk{Scores are higher for \texttt{rand} hay in 19 of 21 LLM--needle combinations, by up to 9.7 points (Gemma, \texttt{neg}).} One could hypothesize that random hay should {\it increase} semantic similarity scores relative to original hay, under the intuition that perturbing a sentence unrelated to the rest of the document does not alter its holistic semantic meaning as significantly. One might likewise expect an increase in scores because original hay provides opportunities for contradictions between the needle sentence and the rest of the document: returning to the example in Figure \ref{fig:d-variants}, 
asserting O'Hare is {\it not} busier than Midway seems at odds with surrounding original context describing it as an airline hub, whereas in a document about the film Goodfellas, whether O'Hare is busier than Midway seems immaterial. 
\pk{The results are consistent with both intuitions, suggesting that} when there is no apparent connection between the needle and its context, LLMs ascribe (on average) \pk{less} semantic importance to perturbations of that needle. We return to this finding in Section \ref{sec:conc}.

Lastly, we consider standard deviation:
for documents with 2 or more sentences, standard deviation \pk{for most LLMs} is decreasing in document length \pk{for both hay types}, suggesting LLMs become more consistent in rating longer documents highly. \pk{The exceptions are GPT-4o and GPT-Luna, whose scores remain noisy at every length.} %

\paragraph{Needle type}

A priori, there is no reason to expect that one perturbation type (negation, named entity replacement, or conjunction swap) changes semantic meaning any more or less than another. Our null hypothesis is equality between all ${3 \choose 2}=3$ pairs of needle types:
\begin{align*}
s(\texttt{neg},H, P) &= s(\texttt{con},H,P) \\
s(\texttt{neg},H, P) &= s(\texttt{ner},H,P) \\
s(\texttt{ner},H, P) &= s(\texttt{con},H,P)
\end{align*}

\pk{We} perform a 2-sample KS test with Bonferroni correction between each of the three comparisons for a given LLM and hay type. We reject the null hypothesis at the highly significant level $(p<0.01)$ in \pk{all 42 settings. The score distributions differ across needle types.}

\begin{figure}[!b]
    \centering
    \includegraphics[width=\linewidth]{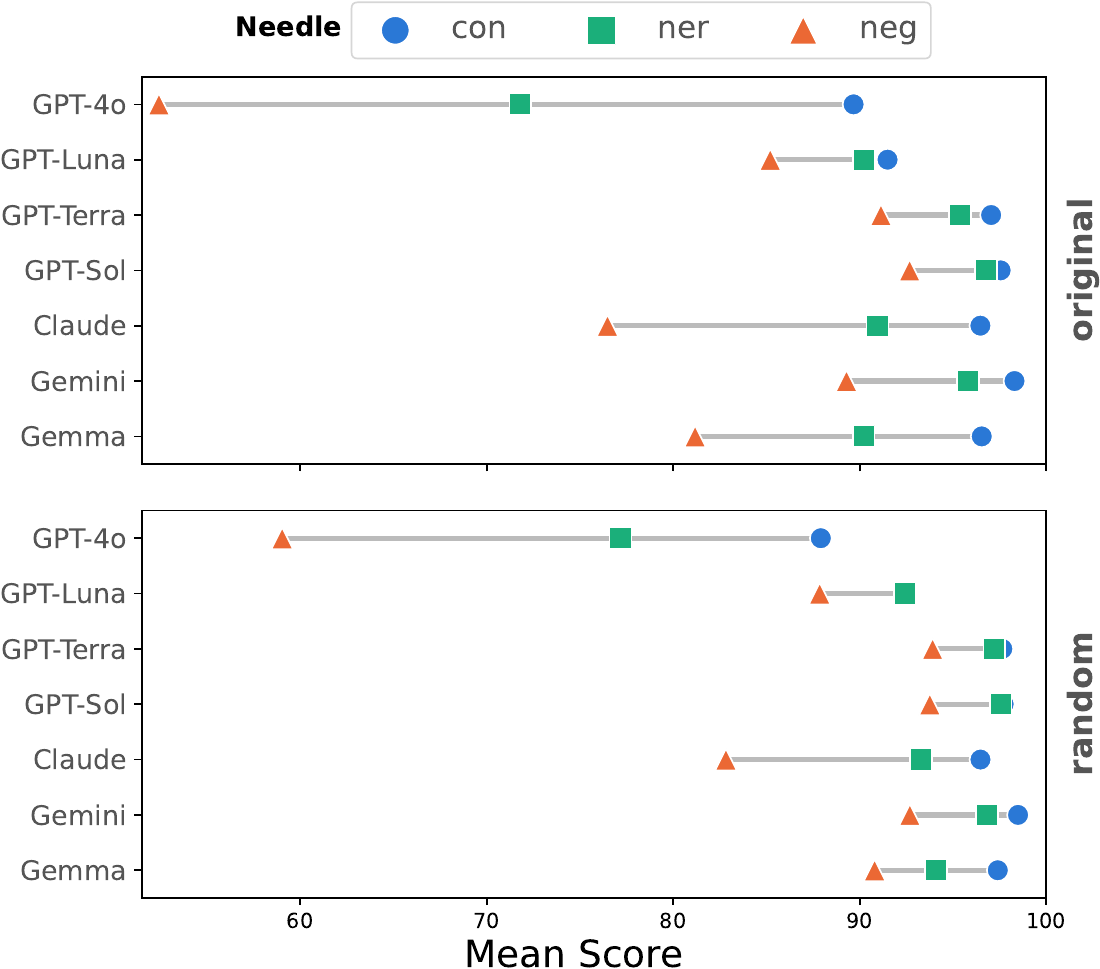}
    \caption{\pk{Mean score by needle type for each LLM under original (top) and random (bottom) hay.}}
    \label{fig:needle_means}
\end{figure}

\pk{Figure~\mbox{\ref{fig:needle_means}} reveals a striking cross-LLM commonality: the ordering of needle types by mean score is the same for every LLM and both hay types,}

\[
\texttt{con} \succ \texttt{ner} \succ \texttt{neg},
\]
\pk{meaning each LLM penalizes negation most, followed by named entity replacement and conjunction swap. One possible interpretation is that the needles differ in how much they change a sentence's claim: negation asserts its opposite, entity replacement keeps the claim but changes whom or what it is about, and a conjunction swap only changes how items or clauses are combined.}

\pk{While the LLMs agree on this ordering, they diverge on how sharply they distinguish the needles. Under original hay, for example, the gap between \texttt{con} and \texttt{neg} is 37 points for GPT-4o, 15 points for Gemma, and only 5 to 6 points for the GPT-5.6 models, which barely distinguish perturbation types. Notably, random hay narrows this gap for every LLM, since it raises \texttt{neg} scores the most (by 1 to 10 points) and \texttt{con} scores the least. Even in the simplest case of single-sentence documents, LLMs disagree widely, scoring a sentence against its negation anywhere from 18 (GPT-4o and Gemma) to 52 (GPT-Sol). In short, LLMs agree on which perturbations matter most, but not on how much.}

\paragraph{Hay type}

Lastly, we study the impact of the surrounding context type -- the hay. \pk{Our analysis so far has shown that all LLMs surveyed judge a perturbation more leniently on average when it is surrounded by unrelated context. Relative to the original context, this random hay not only yields higher scores (Figure~\mbox{\ref{fig:doc_length}}) but also less differentiation between needle types (Figure~\mbox{\ref{fig:needle_means}}) and, for GPT-4o and Gemma, less early-positionality bias (Figure~\mbox{\ref{fig:epb}}). Here, we refine this analysis to the full score distribution, revealing that LLMs differ in \mbox{\textit{which}} scores random hay raises.}

\begin{figure}[t]
    \centering
    \includegraphics[width=\linewidth]{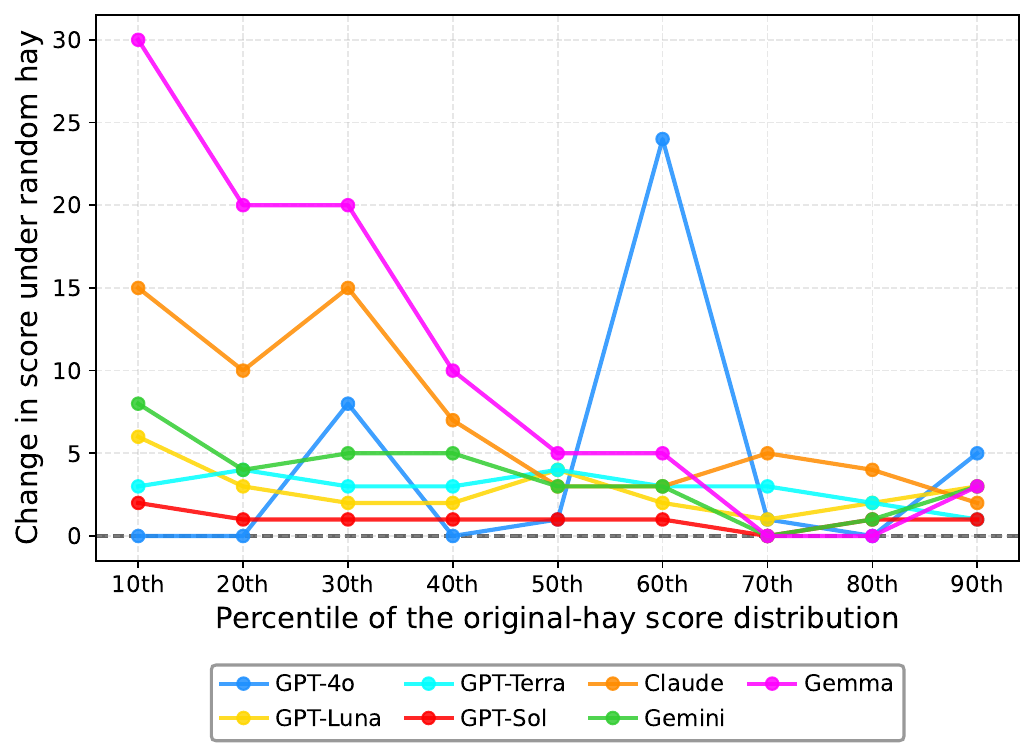}
    \caption{\pk{Change in each decile of the \texttt{neg} score distribution under random hay, relative to original hay (documents with $k \geq 2$).}}
    \label{fig:hay_shift}
\end{figure}

\pk{More precisely, Figure~\mbox{\ref{fig:hay_shift}} plots, for the \texttt{neg} needle, each percentile of the original-hay score distribution against how much that percentile changes under random hay. Gemma and Claude drastically raise the \mbox{\textit{lowest}} scores: their 10th percentile rises by 30 and 15 points, respectively, while scores at the 60th percentile and above rise by 5 points or fewer. In contrast, the GPT-5.6 models raise scores roughly uniformly across the board, by 0 to 6 points at every percentile. Lastly, as evidenced by the sharp spike in Figure~\mbox{\ref{fig:hay_shift}}, GPT-4o's scores change most around the middle of the distribution, where they jump from about 50 under original hay to about 75 under random hay. In this way, while all LLMs here score more leniently under random hay, they exhibit strong differences in which score ranges are most affected.}

\paragraph{Prompt}

Before discussing the effects of varying the prompt, we first acknowledge there is no consensus
definition of document-level semantic similarity. The feasibility and utility of rigorously
defining semantic similarity has been actively researched and debated since at least the 1970s
\citep{Goodman1972,Tversky1977}, with proposals varying vastly in formalism, emphasis,
and according to the textual domain for which they are intended
\citep{Harispe2015,Sogancioglu2017}. For these reasons, the prompt (and choice
of semantic perturbation needles, dataset, value of $k$, etc.) is an experimental design choice
that can be toggled according to one's aims and document corpus.

Regardless of this choice, however, providing {\it some} guidance in the prompt as to what
constitutes semantic similarity is an important guardrail for experiment interpretability,
especially when making cross-LLM comparisons. For the purposes of running our experiment, this
presents a tradeoff: an overly-specified definition narrows experiment generality and risks
artificially constraining the task, while an under-specified one raises concerns that scoring
differences observed simply reflect differences in how LLMs free associate meaning to the phrase
``semantic similarity'' rather than differences in how they {\it judge} (some fixed notion of)
semantic similarity.

The prompt we utilized (see Figure~\ref{fig:criteria}, Appendix~\ref{sec:appendix_params}) aims to
balance this by emphasizing commonalities that persist across definitions of semantic similarity
(such as being \pk{``based on the similarity of underlying semantic concepts and ideas, rather than surface-level lexical
features, word choice, or syntax''}), without adhering to a bespoke definition. Consistent with
suggested best-practices for LLM-as-a-judge \citep{Gu2025}, the
prompt also provides a scoring rubric with anchored scales for each score quartile. In these
regards, our prompt is similar in spirit to that used within the well-known Semantic Textual
Similarity (STS) benchmark \citep{cer2017semeval}, used for scoring semantic similarity
at the sentence-level. To further explore the effects that prompt choice has on our experimental
design and high-level findings, we now consider \pk{three} additional prompts\pk{, each run with Gemma on the same documents (Figure~\mbox{\ref{fig:alt_prompts}}, Appendix~\mbox{\ref{sec:appendix_params}}). We use Gemma for these comparisons because its weights are openly available, so the comparisons can be rerun on the same model, and because, along with GPT-4o, its scores are the most sensitive to several of our parameters, such as needle position and hay type (Figures~\mbox{\ref{fig:epb}} and~\mbox{\ref{fig:doc_length}}). Our aim is not to find the right prompt, but to illustrate what is at stake in this choice, as the prompts approach semantic similarity differently, and with varying degrees of formalism and specificity.}

\begin{figure}[t]
    \centering
    \includegraphics[width=\linewidth]{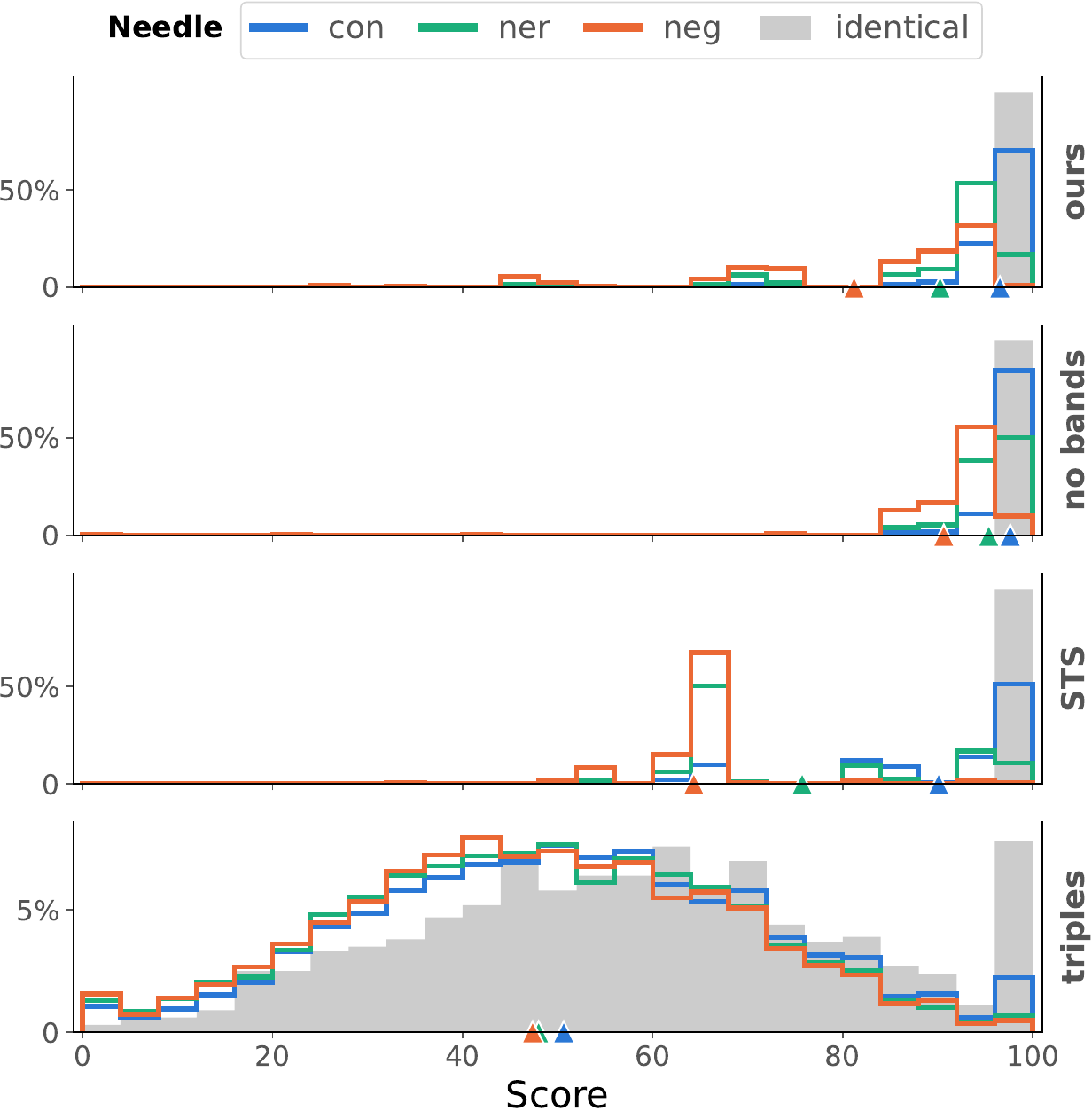}
    \caption{\pk{Gemma's score distributions under four prompts (original hay; the same documents in each panel). Triangles mark needle means, and grey shows scores for identical documents (the null experiment). Note the finer y-axis for triples.}}
    \label{fig:prompts}
\end{figure}

\pk{First, we consider a variant of our prompt where we remove the four scoring bands, leaving only its definition of semantic similarity. Since the rest of the prompt is unchanged, this isolates how much specifying the bands affects output scores.}

\pk{Second}, we adapt the community-standard STS rubric, so that its 0--5 scoring range is
inflated to 0--100. \dev{The two rubrics are similar in form, each anchoring its bands to a
description, but differ in emphasis. Ours attributes more importance and higher scores to topical
overlap, so that a pair of documents sharing ``subject matter, topics, ideas, or context'' scores
high throughout. STS instead descends from equivalence, reserving its top level for documents that
are ``completely equivalent'' and placing ``not equivalent, but are on the same topic'' at level 1
of 5 \citep{cer2017semeval}. Since every pair in our design differs by exactly one sentence, \pk{we expect} the top
STS level \pk{to be} largely foreclosed, whereas ours admits such pairs as ``nearly identical.'' \pk{For these reasons, we}
expect lower scores under STS across the board.}

\pk{Third}, we consider a prompt \dev{built on a fixed mathematical scoring function} \pk{that formalizes and constrains \mbox{\textit{how}} semantic similarity is computed}. In particular,
this prompt asks the LLM to construct a knowledge graph for each document, whose edges are a set of
semantic (subject; relation; object) triples\dev{, in the style of open information extraction \citep{Banko2007}}. Semantic similarity is then scored
using Python as the \dev{Jaccard index} between the edge sets of these graphs using exact match on
canonicalized strings, and then rescaled to the $[0,100]$ range. \pk{In this way, the prompt's approach is akin to computing a distance metric between semantic knowledge graphs.} This prompt differs sharply from
the others: here, the LLM's role is shifted from outputting a score to outputting a structured
representation of semantics, from which a final score is externally and deterministically computed.

\pk{Figure~\mbox{\ref{fig:prompts}} compares the resulting score distributions under original hay. Removing the scoring bands raises scores, increasing the mean \texttt{neg} score from 81 to 91 and halving the gap between \texttt{con} and \texttt{neg} (from 15 to 7 points). The STS rubric has the opposite effect. As anticipated, it lowers scores across the board, with the mean \texttt{neg} score falling to 64 and nearly half of \texttt{neg} pairs placed at exactly 66, the top of its level 3 (``roughly equivalent, but some important information differs or is missing''). Contrary to our expectation, however, its top level is not foreclosed. Although that level is reserved for ``completely equivalent'' documents, Gemma places 75\% of \texttt{con} pairs there, most often at 98.}

\pk{Even so, Gemma's trends from the preceding sections hold under all three prompts in which the LLM outputs a score, namely \mbox{\texttt{con} $\succ$ \texttt{ner} $\succ$ \texttt{neg}}, higher scores under random hay, and higher scores for longer documents. What changes is the \mbox{\textit{magnitude}} of these trends. For instance, Gemma's early-positionality bias for \texttt{neg} drops from 6.0 to 0.9 without the bands. In short, for both the STS and no-bands prompts, we observe changes in how strongly scores respond to each parameter, but not in which direction.}

\pk{The triple prompt behaves very differently. Its scores spread across the whole range, and the mean score by needle type is tightly compressed, differing by only 3 points (47 to 51). This behavior can be explained by a null experiment, the outcome of which is shown in grey in Figure~\mbox{\ref{fig:prompts}}. That is, for each prompt, we check whether comparing a document to itself returns a perfect score, using 10 documents per position and hay type. Under the other three prompts, this is almost always the case, as identical documents score 100 more than 99\% of the time. However, under the triple prompt, identical documents average 58 and score below 50 over a third of the time.}

\pk{The triple prompt fails the null experiment because of stochasticity in the LLM's extraction of triples. For a fixed document, Gemma often extracts different triples in different passes, and exact matching on canonicalized strings then penalizes these differences. For example, the identical sentence ``The half-built sloop was formally christened Stork on 25 May 1756'' scores 33, because in the triple containing the date, one pass uses ``the half-built sloop'' as the subject and the other uses ``Stork.'' As longer documents yield more triples to disagree on, the score of identical documents also falls with length, from 71 for documents of up to three sentences to 53 for those of 14 or more. This noise is so significant that it swamps the effect of the needles: under the triple prompt, a perturbed pair outscores an identical pair of the same length 37 to 41\% of the time, close to the 50\% expected if needles had no effect.}

\pk{We note that strategies exist for minimizing this effect, such as more deterministic decoding (e.g., a temperature of 0), entity disambiguation \mbox{\citep{Hoffart2011}} or clustering of equivalent phrases \mbox{\citep{Galarraga2014}} on the triples, or more lenient matching between triples, such as partial rather than exact matches on canonicalized strings. Regardless, this prompt illustrates how attempts to formalize semantic similarity must be handled with care, as they can create unintended confounds that require additional judgment calls to address.}

\section{Discussion \& Conclusion}\label{sec:conc}

Our results demonstrate that LLM-as-a-judge scoring for pairwise document similarity is sensitive to a range of experimental parameters\pk{, namely needle position, document length, perturbation type, surrounding context, and prompt,} and that these sensitivities manifest differently across models. We now discuss these findings in the context of prior work \pk{and consider what they imply for those who use our framework}, address the scope and limitations of our experimental design, and outline directions for future research.

\paragraph{The effect of context relevance}

\pk{One of our most thought-provoking findings} concerns the effect of context relevance on scoring. \pk{Random hay (surrounding context topically unrelated to the semantic change) \mbox{\textit{raises}} similarity scores.} Two competing intuitions make opposing predictions here. Under a \textit{contradiction view}, original hay permits cross-sentence semantic contradictions that compound the effect of the needle; randomizing the surrounding context removes these contradictions and should yield \textit{higher} similarity scores. Under an \textit{interpretive frame view}, original hay provides a topical context within which the perturbation can be contextualized and even dismissed. For example, a reader encountering one contradictory sentence among nine reinforcing ones might treat it as a minor anomaly rather than a fundamental change in meaning; removing this frame leaves the LLM with no basis on which to downweight the perturbation, predicting \textit{lower} scores under random hay. \pk{Our results favor the contradiction view, as evidenced by the higher mean scores under random hay. Furthermore, that random hay raises scores most for negation (which is plausibly the needle most likely to contradict related sentences) and, for Claude and Gemma, raises their \mbox{\textit{harshest}} original-hay scores the most (Figure~\mbox{\ref{fig:hay_shift}}) are also consistent with the contradiction view.}

\pk{This behavior runs counter to semantic illusions in human comprehension, in which readers and listeners can miss meaning alterations that fit the surrounding discourse} \citep{Cook,Nieuwland}\pk{. For instance, listeners may initially miss that a story's central character has been replaced by an inanimate object \mbox{\citep{Nieuwland}}. LLM judges, by contrast, penalize an alteration more when the surrounding text is related. Although this analogy is limited (e.g., judges compare two documents side by side rather than encountering the change while reading), these} results suggest that LLMs do not simply evaluate the needle in isolation; the coherence of the surrounding context modulates how the perturbation is weighted. For practitioners, this implies that the domain and coherence of compared documents are not neutral factors and actively shape scoring behavior in ways that must be accounted for.

\paragraph{Positionality bias as within-document evaluation weighting}

A growing body of work documents positional biases in LLM-as-a-judge systems, but the existing literature focuses almost exclusively on \textit{candidate-order bias}: when an LLM compares two responses, the order in which they are presented influences the evaluation \citep{wang2023large, zheng2023judging, Shi2024}. Separately, \citet{Liu2023} showed that LLM performance on information retrieval tasks degrades when relevant content appears in the middle of a long context, the so-called ``lost in the middle'' phenomenon. Our findings reveal a complementary and, to our knowledge, previously uncharacterized bias dimension: \textit{within-document positional weighting}, in which the location of a semantic difference within the compared documents themselves influences scoring. 
Unlike the U-shaped retrieval curve of \citet{Liu2023}, our bias concerns \textit{where} within a document a difference occurs and manifests in scoring magnitude rather than retrieval accuracy\pk{. It is detectable for most LLMs, but sizable only for GPT-4o and Gemma, which penalize earlier differences by up to 5 to 6 points, while the remaining judges show little bias and some slightly penalize later differences instead}. Moreover, \pk{for GPT-4o and Gemma, the bias is already present in documents of five or fewer sentences}, well below the long-context regime studied by \citet{Liu2023}, suggesting a distinct mechanism tied to evaluative weighting.

\pk{Crucially, this bias is not a fixed trait of a model. GPT-4o's bias reverses direction for two needle types under random hay, and removing the scoring bands from the prompt nearly eliminates Gemma's bias for \texttt{neg} (from 6.0 to 0.9 points). Whereas candidate-order bias can be exposed by varying the order in which inputs are presented and mitigated by calibration \mbox{\citep{wang2023large}}, the location of a difference is a property of the documents themselves. Within-document positional weighting must therefore be measured for each combination of judge, prompt, and document type, which our factorial design makes straightforward.}

\paragraph{A shared ordering on judge-specific scales}

\pk{All seven LLMs agree on which of our perturbations matter most. For every LLM and hay type, and for Gemma under all four prompts, negation lowers scores the most and conjunction swaps the least. The prominence of negation is notable in light of prior work, in which language models were repeatedly found to treat a sentence and its negation alike} \citep{Kassner2020,Ettinger2020,Truong2023}\pk{, and learned evaluation metrics built on such models underestimated how much a negation changes a sentence's meaning \mbox{\citep{Anschutz2023}}. In contrast, our experiment shows negated single-sentence documents receive mean scores of only 18 to 52 out of 100, the lowest of the three needles for every judge. Because our negation needle always inserts ``not,'' however, our design cannot tell whether judges grasp the meaning of the negated sentence or simply treat any inserted negation as a major change.}

\pk{Despite agreeing on this ordering, the LLMs diverge on the scale on which their scores differentiate the needle types. Under original hay, the gap between \texttt{con} and \texttt{neg} ranges from 5 points for the GPT-5.6 models to 37 for GPT-4o. These differences have practical consequences. Consider, for example, that for 15--19-sentence documents containing a negated sentence, GPT-Sol scores 99\% of pairs at 90 or above, but GPT-4o only 16\%. Consequently, whether such changes would clear a fixed similarity threshold in applications, such as in near-duplicate detection, can depend dramatically on the judge---even for the same vendor. Clearly, model-specific baselines are thus essential for LLM-driven document similarity assessment, with each model version warranting its own characterization.}

\paragraph{The prompt as part of the instrument}

\pk{Our prompt comparison suggests that the prompt is best viewed as part of the measurement instrument rather than a neutral wrapper around it, echoing findings that LLM evaluations can change markedly with instruction wording \mbox{\citep{Mizrahi2024}}. The rubric shapes scores in ways that go beyond rescaling them. GPT-4o's scores cluster at the boundaries of our rubric's bands (Figure~\mbox{\ref{fig:hybrid_heatmap}}), and removing the bands both raises Gemma's scores and nearly removes its positional bias. Unsurprisingly, the prompt's emphases when defining semantic similarity matter as well. Our rubric rewards shared topics, whereas STS rewards equivalence, and STS lowers Gemma's scores most for negation, which preserves a sentence's topic while reversing its claim. This is the distinction between semantic similarity and relatedness \mbox{\citep{Hill2015}}, and the wide spread in how judges score a sentence against its negation suggests that they draw this line differently.}

\pk{Lastly, our triple prompt illustrated how trying to formalize semantic similarity, perhaps in an effort to make it more objective, can nonetheless require new judgment calls to avoid confounding. In particular, the triple prompt replaced the judge's holistic score with a fixed formula, but introduced noise in the triple extraction step, as evidenced by our null experiment. In this sense, it traded ambiguities in how to define and score semantic similarity for new ambiguities in how to define and extract semantic triples from text, for which there are a myriad of differing approaches} \citep{Niklaus2018,Zhou2022}\pk{. It is worth noting that, beyond LLM-driven triple extraction, there are also other graph representations of meaning, such as Abstract Meaning Representation (AMR) graphs \mbox{\citep{Banarescu2013}}, which can be compared with the Smatch metric \mbox{\citep{Cai2013}}, although both are defined for single sentences rather than documents.}

\paragraph{On the idealization of semantic needles}

A natural criticism of our approach is that our chosen perturbation types are idealized interventions that may not reflect the messier reality of natural semantic variation. We respond in two ways.
First, each needle type represents failure modes encountered in real-world, high-stakes settings.
Negation errors arise routinely in clinical documentation, where the presence or absence of ``not'' in a diagnosis or treatment plan is a well-documented patient safety concern, and in regulatory texts where amendments negate previously permitted actions. Conjunctions are \pk{a well-known source of ambiguity} in legal drafting \citep{adams2006revisiting}, as illustrated by the landmark case \textit{O'Connor v.\ Oakhurst Dairy} (2017), in which a missing serial comma and resulting conjunctive ambiguity led to a multi-million-dollar settlement. Named entity replacement occurs in plagiarism and text reuse, template-based document generation, and adversarial text manipulation. %

Second, %
the idealized nature of our needles is a deliberate design choice that confers several advantages. Controlled, atomic perturbations allow us to isolate the effect of individual variables without the confounds introduced by naturalistic variation, where multiple semantic changes co-occur and their effects are entangled. 
The idealization enables automation and scalability: we test \pk{4200} distinct parameter settings across tens of thousands of document pairs, a scale that would be infeasible with hand-crafted naturalistic perturbations. Moreover, because each perturbation is well-defined and reproducible, our framework can be applied to future LLMs consistently, enabling longitudinal comparisons across model versions. 

\paragraph{Limitations}
Our study has several limitations. The corpus is drawn entirely from English Wikipedia, which may not generalize to domain-specific texts (e.g., legal, clinical, or literary documents) where sentence structure and semantic density differ. Our perturbation types, while motivated by real-world scenarios, do not exhaust the space of possible semantic changes; modifications such as quantifier shifts, temporal alterations, or pragmatic implicature changes remain untested. \pk{Our comparison of prompts is limited to one LLM (Gemma) and four prompts, so how other judges respond to changes in prompt wording, scoring scale, or task framing remains untested.} Our analysis treats each LLM as a fixed entity, but model behavior can vary with temperature, system prompt, and API version, and the specific model versions tested here will inevitably be superseded. 
Finally, we do not provide a mechanistic explanation for the observed biases: whether they arise from attention patterns, tokenization effects, or training data artifacts remains an open question.

\paragraph{Future work}

Several directions emerge naturally from our findings. First, our framework's modularity invites extension to new needle types and domains: testing on legal corpora with legally meaningful perturbations, or on clinical texts with dosage and treatment modifications, would assess whether the biases we observe are stable across genres. Second, the pairwise similarity scores produced by our framework can be naturally extended to multi-way comparisons in multiple ways. For example, multi-way rankings can be obtained by constructing win/draw/lose tournaments over collections of documents and analyzing the resulting preference graphs via methods such as Bradley-Terry models. Third, the scoring fingerprints we identify suggest a connection to model architecture and training: systematic comparisons across model families, sizes, and fine-tuning strategies could illuminate which design choices give rise to which scoring behaviors. Finally, because our framework is LLM-agnostic and automated, it can be re-run as new models are released, enabling longitudinal tracking of how scoring behaviors evolve across model versions and providing practitioners with a consistent basis for comparison.

\medskip
\noindent In summary, our needle-in-a-haystack framework provides a scalable, modular, and highly discriminative experimental design for probing LLM scoring behavior in pairwise document similarity. The framework reveals that \pk{LLM judges agree on which semantic changes matter most but express this on judge-specific scales, and that context, position, and prompt shift their scores in judge-specific ways,} phenomena that would be invisible in a simpler experimental setup that varies fewer parameters or aggregates over them. These findings underscore the importance of fine-grained sensitivity testing as a complement to standard benchmarks. We hope the intuitive needle-and-hay framing, together with the quantitative analyses developed here, will serve as a practical toolkit for researchers and practitioners seeking to understand, compare, and audit LLM-as-a-judge document similarity assessment systems for both the \pk{seven} models tested here and for those yet to come. \\

\noindent {\bf Acknowledgements.} We thank Emily Saldanha, Ian Stewart, Joshua Chong, Ana Usenko, and Kate Gibb for helpful conversations and paper feedback.  This work was performed by Pacific Northwest National Laboratory operated by Battelle for the U.S. Department of Energy under Contract DE-AC05-76RL01830. This work was also supported by the Office of the Director of National Intelligence (ODNI), Intelligence Advanced Research Projects Activity (IARPA), via the HIATUS Program contract D2022-2204140001. The views and conclusions contained herein are those of the authors and should not be interpreted as necessarily representing the official policies, either expressed or implied, of ODNI, IARPA, or the U.S. Government. The U.S. Government is authorized to reproduce and distribute reprints for governmental purposes notwithstanding any copyright annotation therein. Information Release:~PNNL-SA-221843.

\subsection*{AI Use Statement}
\pk{The authors retain full responsibility for the paper's content and scientific integrity. In the first version of the paper, generative AI was confined to assisting with figure generation and formatting, improving the clarity and shortening the length of author-written prose, helping search scientific literature for related work, and conducting mock peer reviews. In the second version, AI assistance was again used for these purposes, and additionally to debug the existing author-written Python code for performing the experiment, to draft Python scripts for executing the prompt experiment, and to improve the naturalness of the negation needle, with author verification. Other scientific contributions of the paper, including its conception, the experimental design, and the selection of models to survey, were all developed by the authors, with input from anonymous referees.}

\bibliography{tacl2021}
\bibliographystyle{plainnat}

\appendix

\SetKwComment{Comment}{/* }{ */}

\tikzset{
  process/.style={
    rectangle,
    draw=black,
    fill=blue!20,
    text centered,
    inner sep=6pt, %
  },
  subprocess/.style={
    rectangle,
    draw=black,
    fill=yellow!20,
    text centered,
    inner sep=5pt,
  },
  arrow/.style={
    thick,->,>=stealth
  }
}

\newpage
\begin{figure*}
\centering
\small
\setlength{\fboxsep}{1.5pt}
\begin{tikzpicture}[font=\small]
\node[draw=gray!50, rounded corners=6pt, fill=gray!5, inner sep=5pt] (main) {
    \begin{minipage}{0.97\textwidth}
        \begin{minipage}[t]{\textwidth}
            \centering
            \textbf{Original Document $d$}\\[-2pt]
            {\color{gray!60}\rule{14.2cm}{0.4pt}}\\[2pt]
            \colorbox{cyan!15}{$\dots$}\\[1pt]
            \colorbox{cyan!15}{\parbox{14.2cm}{\centering
                Cleopatra was Queen of the Ptolemaic Kingdom of Egypt.}}\\[1pt]
            \colorbox{yellow!30}{\parbox{14.2cm}{\centering
                A member of the Ptolemaic dynasty, she was a descendant of its founder Ptolemy I Soter.}}\\[1pt]
            \colorbox{cyan!15}{\parbox{14.2cm}{\centering
                After her death, Egypt became a province of the Roman Empire.}}\\[1pt]
            \colorbox{cyan!15}{$\dots$}
        \end{minipage}
        
        \vspace{8pt}
        
        \centering
        \begin{tabular}{c @{\hspace{4mm}} c}
            \textbf{Named Entities} & \textbf{SpaCy Named Entity Tags} \\[-2pt]
            {\color{gray!60}\rule{9.8cm}{0.4pt}} & {\color{gray!60}\rule{4.0cm}{0.4pt}} \\[4pt]
            
            \begin{minipage}{9.8cm}
                \centering
                \colorbox{cyan!15}{\parbox{9.8cm}{\centering \textbf{\textcolor{red}{Cleopatra}} was Queen of the Ptolemaic Kingdom of \textbf{\textcolor{red}{Egypt}}.}}\\[1pt]
                \colorbox{yellow!30}{\parbox{9.8cm}{\centering A member of \textbf{\textcolor{red}{the Ptolemaic dynasty}}, she was a descendant of its founder \textbf{\textcolor{red}{Ptolemy I Soter}}.}}\\[1pt]
                \colorbox{cyan!15}{\parbox{9.8cm}{\centering After her death, \textbf{\textcolor{red}{Egypt}} became a province of \textbf{\textcolor{red}{the Roman Empire}}.}}
            \end{minipage}
            &
            \begin{tikzpicture}[remember picture, baseline=(current bounding box.center)]
                \node[anchor=center, inner sep=0pt, font=\footnotesize] (list) {
                    \begin{minipage}{4.0cm}
                        \centering
                        \tikz[remember picture, baseline]{\node[anchor=base, inner sep=1.5pt] (cleopatra) {\fbox{\textbf{Cleopatra}: PERSON}};}\\[-1pt]
                        \textbf{Egypt}: GPE\\[-1pt]
                        \textbf{the Ptolemaic dynasty}: DATE\\[-1pt]
                        \tikz[remember picture, baseline]{\node[anchor=base, inner sep=1.5pt] (ptolemy) {\fbox{\textbf{Ptolemy I Soter}: PERSON}};}\\[-1pt]
                        \textbf{the Roman Empire}: GPE
                    \end{minipage}
                };
                \draw[->, thick, red!70, bend right=60] 
                (cleopatra.west) to node[midway, sloped, above]{\small replace} (ptolemy.west);
            \end{tikzpicture}
        \end{tabular}
        
        \vspace{8pt}
        
        \begin{minipage}[t]{\textwidth}
            \centering
            \textbf{Document with Named Entity Replacement $\hat d$}\\[-2pt]
            {\color{gray!60}\rule{14.2cm}{0.4pt}}\\[2pt]
            \colorbox{cyan!15}{$\dots$}\\[1pt]
            \colorbox{cyan!15}{\parbox{14.2cm}{\centering
                Cleopatra was Queen of the Ptolemaic Kingdom of Egypt.}}\\[1pt]
            \colorbox{yellow!30}{\parbox{14.2cm}{\centering
                A member of the Ptolemaic dynasty, she was a descendant of its founder \textbf{\textcolor{red}{Cleopatra}}.}}\\[1pt]
            \colorbox{cyan!15}{\parbox{14.2cm}{\centering
                After her death, Egypt became a province of the Roman Empire.}}\\[1pt]
            \colorbox{cyan!15}{$\dots$}
        \end{minipage}
    \end{minipage}
};
\end{tikzpicture}
\vspace{-5pt}
\caption{Named entity replacement needle: spaCy NER identifies entities in the selected sentence; one is chosen via random sampling and its span is replaced with the text of a randomly chosen entity of the same label drawn from elsewhere in the document.}
\label{fig:ner-example}
\end{figure*}

\section{Experimental Design Details}
\label{sec:appendix_params}

\paragraph{Data selection, cleaning, \& length}

The \texttt{Plain Text Wikipedia} documents used in our experiments are publicly available through Kaggle \cite{plain_text_wikipedia_2020}. Though these documents are already processed as plain text, they are further cleaned to remove all section headers, appendical sections (e.g.~``See Also'', ``References''), and other non-natural language text such as tables and code. We parse sentences using spaCy and only consider documents with \pk{\mbox{$L \geq 30$}} sentences, discarding any documents with fewer sentences or with average sentence length that is unusually short or long, which we define as fewer or more than $150L$ and $2000L$ characters, respectively.

We vary our position parameters $0 \leq i, j \leq 9$, which yields cleaned documents ranging from 1 sentence to 19 sentences and spans 100 distinct needle positions. The maximum $i,j$ value of $9$ was selected to strike a balance between granularity and length, keeping documents within typical LLM context windows while providing enough positional variation to detect biases. %

\paragraph{Needle type implementations}

We implement the negation (\texttt{neg}), conjunction swap (\texttt{con}) and named-entity replacement (\texttt{ner}) needle types in Python using spaCy v3.8.11 under model \texttt{en\_core\_web\_sm} for tokenization, POS tagging, dependency parsing, and NER \cite{spacy}. 

\begin{itemize}
    \item \texttt{neg}: for each sentence marked for negation perturbation, we first detect existing negation via the ``neg'' dependency or tokens with lemmas not/n’t/never/no; if present, we skip the sentence and move to the next. \pk{Otherwise, we insert sentential negation at the sentence's root verb: if the root is a copula or auxiliary, we insert ``not'' after it (``was'' \mbox{$\to$} ``was not''); if the root verb has an auxiliary, we insert ``not'' after the first auxiliary (``has worked'' \mbox{$\to$} ``has not worked''); otherwise, we add do-support (``published'' \mbox{$\to$} ``did not publish''). Sentences without a unique verbal root that has a subject are skipped.} For example, in the sentence:
 \begin{center}
\setlength{\tabcolsep}{4pt} 

\small 

\begin{tabular}{ccccc}
Caesar & \textcolor{red}{was} & a & Roman & general. \\[1ex]

\textcolor{gray}{$\downarrow$} & \textcolor{gray}{$\downarrow$} & \textcolor{gray}{$\downarrow$} & \textcolor{gray}{$\downarrow$} & \textcolor{gray}{$\downarrow$} \\[1ex]

\textcolor{gray}{nsubj} & \textbf{ROOT} & \textcolor{gray}{det} & \textcolor{gray}{amod} & \textcolor{gray}{attr}
\end{tabular}
\end{center}
spaCy dependency parsing identifies the root\pk{, a copula,} and inserts ``not'' immediately after it, yielding: 
\begin{center}
\small
Caesar was \textcolor{red}{not} a Roman general.
\end{center}
    We preserve casing, spacing, and punctuation and avoid contractions to prevent double negatives.
    \item \texttt{con}: we swap occurrences of ``and'' with occurrences of ``or'' in a sentence, and vice versa using spaCy's POS tagging, searching the text for ``and'' and ``or'' and swapping accordingly, as illustrated before in Figure \ref{fig:d-variants}. We preserve original casing and do not alter ampersands.
    \item \texttt{ner}: for each document, we inspect the middle sentence and, if it contains an eligible entity (label $\in$ {\texttt{PERSON, GPE, LOC, LANGUAGE, DATE}}), we select one uniformly at random and replace its text span with the text of a randomly chosen entity of the same label from elsewhere in the document. Figure \ref{fig:ner-example} presents an example\footnote{Note that in Figure \ref{fig:ner-example} spaCy's model incorrectly labels ‘the Ptolemaic dynasty’ as \texttt{DATE}. While such errors are inevitable in natural language processing and contribute noise to the NER needle, this is partially mitigated by our large sample size and aggregate analysis over many documents. }. Replacements operate on the exact span returned by spaCy, treating multi-token entities as a unit and preserving surrounding whitespace and punctuation. %
\end{itemize}

\paragraph{Handling failed semantic perturbations} In some cases, this middle sentence cannot be semantically altered as desired. For example, the middle sentence might have no eligible entity, or the document lacks another entity of the same label to substitute, rendering \texttt{ner} impossible. Similarly, a sentence without connectives ``and/or'' cannot be altered via \texttt{con}. 

In such cases, we check $5$ sentences above and below for sentences that can be satisfactorily altered. Here, recall our minimum document length of \pk{30} sentences, along with our chosen maximum position $(i,j)=(9,9)$, \pk{which yields windows of at most 19 sentences}. This guarantees this scanning procedure never places the needle in positions such that the end of the document would be reached when appending varying amounts of hay. \pk{If no such sentence is found, the document is discarded in favor of the next document from \mbox{$\mathcal{C}$}. Because a document may admit one needle type but not another, screening each needle type independently would compare needle types on different documents. We therefore pre-screen \mbox{$\mathcal{C}$} once and retain only documents admitting all three perturbations, so that each position \mbox{$(i,j)$} is scored on an identical document set across needle types. Screening uses spaCy alone and requires no model calls; it retains 92\% of the documents screened.}

\paragraph{Random hay implementation} 
Documents with random hay (\texttt{rand}) retain their needle sentence, but surround it with sentences from a randomly chosen document. More precisely, for a document $d(N,\texttt{rand}, (i,j))$ whose $m$'th sentence was selected as the needle, a randomly chosen document from $\mathcal{C}$ is selected, and we extract a consecutive window of $i$ sentences before and $j$ sentences after its $m$'th sentence. This random context is then placed around the needle sentence, as shown in Figure \ref{fig:d-variants}. This random selection is made consistently across LLMs, guaranteeing $d(N,\texttt{rand},(i,j))$ is the same across all choices of LLM $L$. \pk{More generally, every LLM scores the same documents, except where an LLM declined or failed to score a document (e.g., because of a provider's content filter), in which case the next document was used instead; this affected 695 documents for GPT-4o, 37 for Claude, and 15 for Gemini.}

\begin{figure}[t]
\centering
\footnotesize
\setlength{\fboxsep}{1.5pt}
\begin{tikzpicture}[font=\footnotesize]
\node[draw=gray!50, rounded corners=6pt, fill=gray!5, inner sep=5pt] (main) {
    \begin{minipage}{0.45\textwidth}
        \centering
        \textbf{Semantic Similarity}\\[-2pt]
        \textbf{(Scale: 0--100)}\\[-2pt]
        {\color{gray!60}\rule{0.7\textwidth}{0.4pt}}\\[6pt]
        
        \parbox{0.95\linewidth}{\textit{Please score the degree to which two text documents are semantically similar. Semantic similarity is based on the similarity of underlying semantic concepts and ideas, rather than surface-level lexical features, word choice, or syntax.}}
        \\[6pt]
        
        \colorbox{red!20}{\parbox{0.95\linewidth}{\textbf{0--25:} The documents have significantly different semantic meaning. The documents convey fundamentally different subject matter, topics, ideas, or context.}}
        \\[2pt]
        
        \colorbox{orange!20}{\parbox{0.95\linewidth}{\textbf{26--50:} The documents share some overlap in semantic meaning, but semantic differences outweigh similarities. There are significant differences in the subject matter, topic, ideas, or context.}}
        \\[2pt]
        
        \colorbox{yellow!20}{\parbox{0.95\linewidth}{\textbf{51--75:} The documents share a substantial amount of semantic similarity. Semantic similarities outweigh differences, but there are still nontrivial ways in which the subject matter, topics, ideas or context differ.}}
        \\[2pt]
        
        \colorbox{green!20}{\parbox{0.95\linewidth}{\textbf{76--100:} The documents have nearly identical or identical semantic meaning. Both documents convey the same core idea and information and the same subject matter, topics, idea, and context.}}
        
    \end{minipage}
};
\end{tikzpicture}
\caption{Criteria: Scoring rubric for evaluating semantic similarity between documents\pk{ (verbatim, as sent to each LLM)}.}
\label{fig:criteria}
\end{figure}

\begin{figure*}[t]
\centering
\footnotesize
\setlength{\fboxsep}{1.5pt}
\newlength{\altLw}\setlength{\altLw}{0.610\textwidth}\newlength{\altRw}\setlength{\altRw}{0.370\textwidth}
\newlength{\altLi}\setlength{\altLi}{0.577\textwidth}\newlength{\altRi}\setlength{\altRi}{0.337\textwidth}
\begin{minipage}[t]{\altLw}
\centering
\begin{tikzpicture}[font=\footnotesize, baseline=(current bounding box.north)]
\node[draw=gray!50, rounded corners=6pt, fill=gray!5, inner sep=4pt] {
    \begin{minipage}[t]{\altLi}
        \centering
        \textbf{No bands}\\[-2pt]
        \textbf{(Scale: 0--100)}\\[-2pt]
        {\color{gray!60}\rule{0.7\textwidth}{0.4pt}}\\[3pt]
        \parbox{0.95\linewidth}{\textit{Please score the degree to which two text documents are semantically similar. Semantic similarity is based on the similarity of underlying semantic concepts and ideas, rather than surface-level lexical features, word choice, or syntax.}}
        \\[3pt]
        \colorbox{gray!15}{\parbox{0.95\linewidth}{\textbf{0--100:} Output a score between 0 and 100 according to the criteria.}}
    \end{minipage}
};
\end{tikzpicture}\\[4pt]
\begin{tikzpicture}[font=\footnotesize, baseline=(current bounding box.north)]
\node[draw=gray!50, rounded corners=6pt, fill=gray!5, inner sep=4pt] {
    \begin{minipage}[t]{\altLi}
        \centering
        \textbf{Triples}\\[-2pt]
        \textbf{(one document at a time)}\\[-2pt]
        {\color{gray!60}\rule{0.7\textwidth}{0.4pt}}\\[3pt]
        \parbox{0.95\linewidth}{\raggedright Extract the relational triples asserted by the document below.\\[4pt]
        A triple has the form (subject; relation; object). The subject and the object are the two things being related. The relation is the phrase in the text that links them; it need not be a verb, and there is no fixed vocabulary of relations. Extract every relation the document asserts. Do not extract anything the document does not assert.\\[4pt]
        Return a JSON array and nothing else, in the form\\
        \hspace*{1em}\texttt{[["subject", "relation", "object"], ...]}\\[4pt]
        Document:\\
        \textit{[document text]}}
    \end{minipage}
};
\end{tikzpicture}
\end{minipage}\hfill
\begin{minipage}[t]{\altRw}
\centering
\begin{tikzpicture}[font=\footnotesize, baseline=(current bounding box.north)]
\node[draw=gray!50, rounded corners=6pt, fill=gray!5, inner sep=4pt] {
    \begin{minipage}[t]{\altRi}
        \centering
        \textbf{STS}\\[-2pt]
        \textbf{(Scale: 0--100)}\\[-2pt]
        {\color{gray!60}\rule{0.7\textwidth}{0.4pt}}\\[3pt]
        \parbox{0.95\linewidth}{\textit{Score the degree to which two text documents are semantically similar, using the following six-level scale.}}
        \\[3pt]
        \colorbox{red!20}{\parbox{0.95\linewidth}{\textbf{0--16 (Level 0):} The two documents are completely dissimilar.}}\\[2pt]
        \colorbox{orange!20}{\parbox{0.95\linewidth}{\textbf{17--33 (Level 1):} The two documents are not equivalent, but are on the same topic.}}\\[2pt]
        \colorbox{yellow!20}{\parbox{0.95\linewidth}{\textbf{34--49 (Level 2):} The two documents are not equivalent, but share some details.}}\\[2pt]
        \colorbox{lime!20}{\parbox{0.95\linewidth}{\textbf{50--66 (Level 3):} The two documents are roughly equivalent, but some important information differs or is missing.}}\\[2pt]
        \colorbox{green!15}{\parbox{0.95\linewidth}{\textbf{67--83 (Level 4):} The two documents are mostly equivalent, but some unimportant details differ.}}\\[2pt]
        \colorbox{green!30}{\parbox{0.95\linewidth}{\textbf{84--100 (Level 5):} The two documents are completely equivalent, as they mean the same thing.}}
    \end{minipage}
};
\end{tikzpicture}
\end{minipage}
\caption{\pk{The three alternative prompts of the Prompt paragraph in Section~\mbox{\ref{sec:results}}. The no-bands and STS prompts replace the criteria block of Figure~\mbox{\ref{fig:criteria}}; the STS levels are adapted from \mbox{\citet{cer2017semeval}}, with ``sentences'' replaced by ``documents.'' The triple prompt is sent once per document, and the score is computed from the two responses in Python.}}
\label{fig:alt_prompts}
\end{figure*}

\paragraph{Prompt, scoring rubric, and trial independence} 
Figure \ref{fig:criteria} presents the prompt and scoring rubric we utilize in our experiment.
Semantic similarity has been defined in many ways—mathematically, algorithmically, and linguistically—and the literature offers robust frameworks for doing so \cite{Harispe2015, Dekang1998}. In our prompt, we adopt a deliberately broad, intuitive notion that emphasizes shared underlying meaning across words, phrases, and sentences, without committing to a specific formal definition. This choice allows us to assess LLM-as-a-judge’s ability to recognize semantic similarity based on widely held intuitions, rather than its adherence to any particular formalism. 

Lastly, it is important to note we use stateless API calls when accessing the LLMs to ensure each prompt is processed in isolation. This ensures that, for a given LLM, each scoring trial is independent of the next, with no shared context window. \pk{Needle randomization draws from a dedicated pseudorandom generator seeded from the experiment configuration, so the perturbed documents are reproducible from that seed alone, while each judge call receives its own random seed where the API supports one.}

\paragraph{Alternative prompts}
\pk{Figure~\mbox{\ref{fig:alt_prompts}} gives the three alternative prompts of Section~\mbox{\ref{sec:results}}, each run with Gemma 4 on the same documents, positions, and needles as our main experiment. The no-bands and STS prompts keep the instructions surrounding the criteria block. For the triple prompt, each extracted field is lowercased and stripped of accents, punctuation, and leading articles, and each relation is lemmatized with spaCy \mbox{\citep{spacy}}. The score is then \mbox{$100\,|A\cap B|/|A\cup B|$}, where $A$ and $B$ are the two documents' sets of triples, matched exactly. The 0.8\% of document pairs whose responses cannot be parsed are discarded and replaced by the next document. In the null experiment, each of the four prompts scores 10 documents per position against themselves, for each hay type.}

\paragraph{Choosing number of documents}

For a given position $(i,j)$, we continue scoring documents until two criteria are met: (1) at least $N$ documents have been processed; and (2) maximum difference in the running mean over the last $w$ documents is less than given threshold, $t$. More formally, $\text{nDoc}(i,j)$, the number of documents we process for position $(i,j)$, is defined by
\[
\min\Bigl\{\, n \ge N :
\max_{\substack{a,b \in \{n-w,\dots,n\}}}
\bigl|\overline{s}_a - \overline{s}_b\bigr| \le t
\,\Bigr\}.
\]
where $\overline{s_a}$ denotes the mean score over documents $1,\dots, a$. \pk{Here \mbox{$n$} counts documents for which the judge returned a score; calls that fail are retried, and a document that cannot be scored is replaced rather than counted.} We choose $N=100$, $w=10$, and $t=1$. After all positions have been processed (for a fixed LLM, needle, and hay type), we take $D=\max_{i,j}\mbox{nDoc}(i,j)$, loop back over any positions for which fewer documents have been processed, and continue until $\mbox{nDoc}(i,j)=D$ for all $i,j$. We note that the 100 document minimum ends up being close to sufficient for satisfying the second criterion for many positions $(i,j)$: across all positions, needles, and LLMs, we see nDoc ranging \pk{from 100 to 111 for original hay and from 100 to 113 for random hay}. 

\begin{table}[!htb]
\footnotesize
\centering
\caption{Details of the Language Models Used}
\label{table:llm-details}
\begin{tabular}{|l|p{2.7cm}|p{1.6cm}|}
\hline
\textbf{Model} & \textbf{Version}  & \textbf{Host} \\ 
\hline
GPT-4o            & 2024-11-20 & Azure    \\ \hline
GPT 5.6 Sol       & 5.6        & Azure    \\ \hline
GPT 5.6 Luna      & 5.6        & Azure    \\ \hline
GPT 5.6 Terra     & 5.6        & Azure    \\ \hline
Claude Sonnet 5   & 5          & AWS      \\ \hline
Gemini 3.7 Flash  & gemini-3.7-flash & GCP \newline Vertex AI \\ \hline
Gemma 4           & google-gemma-4-26b-a4b-it & Azure \\ \hline
\end{tabular}
\end{table}

\paragraph{LLM Information \& Availability}

All Large Language Models (LLMs) were accessed via APIs and used with default hyperparameter settings, including default values for temperature, top-$p$, max tokens, and other generation parameters as provided by their respective APIs. \pk{The Azure OpenAI API was used to access GPT-4o and Gemma 4, and a private API was used to access Gemini 3.7 Flash, Claude Sonnet 5, GPT 5.6 Sol, GPT 5.6 Luna, and GPT 5.6 Terra,} see Table \ref{table:llm-details}. Our experimental code may be made available upon request, pending institutional and sponsor approval.

\paragraph{Score stability}
\pk{Each call uses the API's default temperature and, where the API supports one, its own random seed, so the same document pair can receive different scores in different calls. Our design contains a direct check of this variability: position \mbox{$(0,0)$} has no hay, so for \texttt{neg} and \texttt{con}, whose perturbations are deterministic, its original-hay and random-hay runs score identical inputs independently. Across LLMs, the two runs give exactly the same score for 19\% to 70\% of these pairs, and differ by 2.5 to 10.8 points on average. This check is limited to single-sentence documents. We address the potential noise of individual runs in two ways: (1) via our aforementioned stopping criteria, under which we score each position until its running mean changes by at most one point over the last ten documents (and on no fewer than 100 documents); and (2) by reporting tests and confidence intervals when possible. Lastly, we did not vary temperature or other generation settings, which can have significant effects; we leave this to future work.}

\end{document}